\documentclass[manuscript,screen]{acmart}

\usepackage{times}
\usepackage{epsfig}
\usepackage{graphicx}
\usepackage{amsmath}
\usepackage[normalem]{ulem}
\usepackage{float} 
\usepackage{stfloats}

\usepackage{xcolor} 
\usepackage[flushleft]{threeparttable} 
\usepackage{multirow} 
\usepackage{soul} 
\usepackage{wrapfig,lipsum,booktabs} 
\usepackage{pifont}
\usepackage{comment}

\newcommand{\PaperTitle}{DANCE: DAta-Network Co-optimization for Efficient Segmentation Model Training and Inference}
\newcommand{\FrameworkName}{DANCE}
\definecolor{Note_color}{rgb}{0.0, 0.0, 0.0}
\newcommand{\NOTE}[1]{\textcolor{Note_color}{#1}}
\definecolor{TBD_color}{rgb}{1.0, 0.0, 0.0}

\newcommand{\TrainingEnergySaving}{$\downarrow$25\% - $\downarrow$77\%}
\newcommand{\InferenceEnergySaving}{$\downarrow$31\% - $\downarrow$56\%}
\newcommand{\mIoUBoosting}{$\downarrow$0.71\% - $\uparrow$ 13.34\%}

\AtBeginDocument{%
  \providecommand\BibTeX{{%
    \normalfont B\kern-0.5em{\scshape i\kern-0.25em b}\kern-0.8em\TeX}}}

\setcopyright{acmcopyright}
\copyrightyear{2021}
\acmYear{2021}
\acmDOI{TBD}

\acmJournal{TODAES}
\acmVolume{0}
\acmNumber{0}
\acmArticle{0}
\acmMonth{0}

\begin{document}

\title{\PaperTitle}


{\let\thefootnote\relax\footnote{{This work was supported by the National Science Foundation through the real-time machine learning (RTML) program (Award number: 1937592).}}}

\author{Chaojian Li}
\affiliation{%
  \institution{Rice University}
  \streetaddress{6100 Main ST}
  \city{Houston}
  \country{USA}}
\email{cl114@rice.edu}

\author{Wuyang Chen}
\affiliation{%
  \institution{University of Texas at Austin}
  \streetaddress{110 Inner Campus DR}
  \city{Austin}
  \country{USA}}
\email{wuyang.chen@utexas.edu}

\author{Yuchen Gu}
\affiliation{%
  \institution{Rice University}
  \streetaddress{6100 Main ST}
  \city{Houston}
  \country{USA}}
\email{yg50@rice.edu}

\author{Tianlong Chen}
\affiliation{%
  \institution{University of Texas at Austin}
  \streetaddress{110 Inner Campus DR}
  \city{Austin}
  \country{USA}}
\email{tianlong.chen@utexas.edu}

\author{Yonggan Fu}
\affiliation{%
  \institution{Rice University}
  \streetaddress{6100 Main ST}
  \city{Houston}
  \country{USA}}
\email{yf22@rice.edu}

\author{Zhangyang Wang}
\affiliation{%
  \institution{University of Texas at Austin}
  \streetaddress{110 Inner Campus DR}
  \city{Austin}
  \country{USA}}
\email{atlaswang@utexas.edu}

\author{Yingyan Lin}
\affiliation{%
  \institution{Rice University}
  \streetaddress{6100 Main ST}
  \city{Houston}
  \country{USA}}
\email{yingyan.lin@rice.edu}

\renewcommand{\shortauthors}{Li, et al.}

\begin{abstract}
Semantic segmentation for scene understanding is nowadays widely demanded, raising significant challenges for the algorithm efficiency, especially its applications on resource-limited platforms.
Current segmentation models are trained and evaluated on massive high-resolution scene images (``data level'') and suffer from the expensive computation arising from the required multi-scale aggregation (``\NOTE{network} level'').
In both folds, the computational and energy costs in training and inference are notable due to the often desired large input resolutions and heavy computational burden of segmentation models.
To this end, we propose DANCE, general automated \textbf{DA}ta-\textbf{N}etwork \textbf{C}o-optimization for \textbf{E}fficient segmentation model \textbf{training and inference}. Distinct from existing efficient segmentation approaches that focus merely on light-weight \NOTE{network} design, DANCE distinguishes itself as an automated \textbf{simultaneous} data-\NOTE{network} \textbf{co-optimization} via both input data manipulation and \NOTE{network} architecture slimming. Specifically, DANCE integrates automated data slimming which adaptively downsamples/drops input images and controls their corresponding contribution to the training loss guided by the images' spatial complexity. Such a downsampling operation, in addition to slimming down the cost associated with the input size directly, also shrinks the dynamic range of input object and context scales, therefore motivating us to also adaptively slim the \NOTE{network} to match the downsampled data. Extensive experiments and ablating studies (on four SOTA segmentation models with three popular segmentation datasets under two training settings) demonstrate that DANCE can achieve \textbf{``all-win''} towards efficient segmentation (reduced training cost, less expensive inference, and better mean Intersection-over-Union (mIoU)).
Specifically, DANCE can reduce \TrainingEnergySaving~energy consumption in training, \InferenceEnergySaving~in inference, while boosting the mIoU by \mIoUBoosting.
\end{abstract}

\ccsdesc[500]{Computing methodologies~Image segmentation}
\ccsdesc[300]{Computing methodologies~Neural networks}
\ccsdesc[300]{Hardware~Platform power issues}

\keywords{efficient training and inference methods, semantic segmentation}

\maketitle

\section{Introduction}

\label{sec:intro}
The recent record-breaking performance of semantic segmentation using deep networks motivates an ever-growing application demand. However, those segmentation models typically bear a heavy computational cost to run (i.e., \textbf{inference}), making them extremely challenging to be deployed into resource-constrained platforms, ranging from mobile phones to wearable glasses, drones, and autonomous vehicles. Particularly, while existing works on improving inference efficiency are traditionally focused on classification, state-of-the-art (SOTA) segmentation models are even much more costly. For example, a ResNet50~\cite{resnet} costs 4 GFLOPs for inference with an input size $224 \times 224$. In comparison, for a DeepLabv3+~\cite{deeplabv3+} with the Resnet50 backbone and the same $224 \times 224$ input (associated with an output stride  of 16), the inference cost jumps up to 13.3 GFLOPs; the cost could further soar to 435 GFLOPs if we operate on a higher input resolution of $2048 \times 1024$. A similar trend can be expected in terms of the required energy costs.
These highly required resource costs prohibit segmentation models from edge device deployments or at least degrade the quality of user experience. Specifically, such expensiveness of segmentation models arises from two aspects:

\begin{itemize}
    \setlength{\itemsep}{0pt}
    \setlength{\parsep}{0pt}
    \item \textit{High input resolution and its proportional costs: } segmentation, as a dense prediction task, typically relies on fully convolutional \NOTE{networks} whose inference FLOPs are proportional to the input size. Meanwhile, unlike classification, segmentation is well-known to be more resolution-sensitive due to its much finer prediction granularity~\cite{chen2019collaborative}. Therefore, high-resolution inputs are preferable for improving algorithmic performance, which yet contradicts the resource-saving needs. 
    \item \textit{Multi-scale aggregation: } segmentation is well-known for its strong dependency on multiple scale features~\cite{deeplabv3+,yu2018bisenet,yu2015multi,zhao2018icnet,zhao2017pyramid} for contextual reasoning in combination with full-resolution outputs. Such a desired feature is often achieved by fusing a multi-resolution stream or aggregating paralleled filters with different sizes. Both the fusion and aggregation modules can incur heavy resource costs.
\end{itemize}

The expensiveness of segmentation is further amplified when we come to consider its \textbf{training} (e.g., continuous learning and adaptation) in resource-constrained settings. Many applications, such as autonomous vehicles and robots, require real-time and in-situ learning and continuous adaptation to new data, to be considered truly intelligent. As compared to cloud-based (re)training, local (re)training helps avoid transferring data back and forth between data centers and local platforms, reducing communication loads, and enhancing privacy. Besides, the increasingly prohibitive energy, financial and environmental costs of training ML algorithms have become a growing concern even for training in the cloud~\cite{strubell2019energy}. However, resource-constrained training was not explored much until a few recent efforts on classification~\cite{biggest_loser,e^2_train,you2019drawing}. 

\textbf{Our contributions.} This work aims to push forward the training and inference efficiency of SOTA segmentation models to a new level, from the current practice of merely focusing on light-weight \NOTE{network design}, towards \textbf{a novel data-\NOTE{network} co-optimization perspective}. Its core driving motivation can be summarized in \textbf{two points}: (1) \textit{not all input samples are born equal}~\cite{biggest_loser,Li_2017}; and (2) \textit{eliminating input variances reduces the model's learning workload}~\cite{engelbrecht1999variance}.

More specifically, we propose \FrameworkName, an efficient training and inference framework that can be applied towards any existing segmentation model. First, \FrameworkName~adopts an input adaptive automated data slimming technique. We propose a spatial complexity indicator to adapt the input images' spatial resolution, training sampling frequency, and weighted coefficients in the loss function. Thus \FrameworkName~makes the models focus more on the complicated samples during training, while during testing the input images' spatial resolution will be similarly reduced (i.e., downsampled). 

Meanwhile, adaptively reducing input resolution has direct (proportional) impacts on the training and inference energy costs (i.e., both computation and memory movement costs). The indirect, yet also the important consequence is that the downsampled inputs become more ``normalized'' in terms of object and feature scales. Current segmentation models strongly rely on built-in multi-scale aggregation modules, to balance between contextual reasoning and fine-detail preservation~\cite{chen2019collaborative,yu2015multi}. Interestingly, with spatial-complexity-adaptive downsampled inputs, further slimming those cost-dominant multi-scale aggregation building blocks save both training and inference costs without hampering the algorithmic performance, that's our proposed automated \NOTE{network} slimming in \FrameworkName.

\begin{figure}[t]
   \centering
   \includegraphics[width=0.9\linewidth]{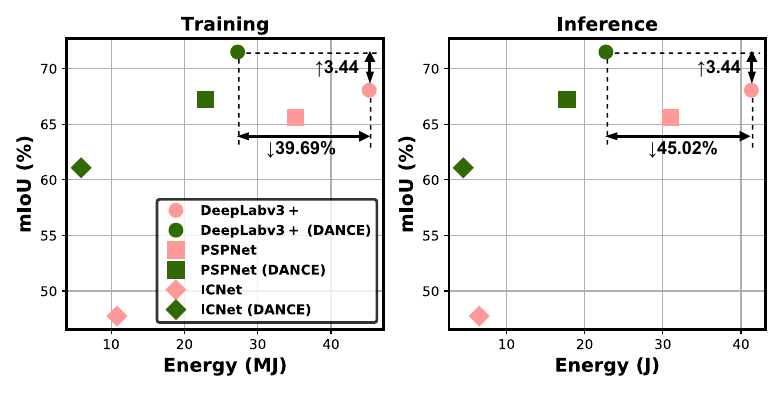}
    \vspace{-1em}
   \caption{The achieved mIoU vs. the required energy cost (Left: the total training energy cost; Right: the averaged inference energy cost per image) on the Cityscapes~\cite{cordts2016cityscapes} test dataset. For the three segmentation models evaluated, \FrameworkName~achieves ``all-win'': reduced training cost, less expensive inference, and improved mIoU.}
   \vspace{-1em}
\label{fig:acc_energy}
\end{figure}

Below we outline the contributions of the proposed \FrameworkName~framework:
\begin{itemize}
    \item \FrameworkName, \textbf{the first} data-\NOTE{network} co-optimization framework, boosts efficiency of both training and inference for segmentation models while mostly improving the accuracy. Further, \FrameworkName~is general and thus can be applied to any existing segmentation backbone.
    \item \FrameworkName~in this paper simultaneously integrates automated data and \NOTE{network} slimming to manipulate input images and their contribution to the model while slimming the \NOTE{network} architecture in a co-optimization manner. Interestingly, the former can emulate the effect of multi-scale aggregation, thus enabling more aggressive slimming of their corresponding cost-dominant building blocks.
    \item Extensive experiments and ablation studies demonstrate that \FrameworkName~can achieve \textbf{``all-win''} (i.e., reduced training and inference costs, and improved model accuracy) towards efficient segmentation, when benchmarking  on four SOTA segmentation models and three popular segmentation benchmark datasets. As shown in Fig.~\ref{fig:acc_energy}, \FrameworkName~establishes \textbf{a new record trade-off} between segmentation models' accuracy and training\&inference efficiency.
\end{itemize}

\section{Related works}
\subsection{Efficient CNN inference and training}
Extensive works have been proposed to improve the efficiency of CNN inference, most of them focus on the classification tasks. \NOTE{Network} compression has been widely studied to speed up CNN inference, e.g., by pruning unimportant \NOTE{network} weights~\cite{han2015deep,he2017channel}, quantizing the \NOTE{network} into low bitwidths~\cite{hubara2017quantized}, or distilling lighter-weight \NOTE{networks} from teachers~\cite{polino2018model}. For example, a representative automated pruning method (Network Slimming~\cite{liu2017learning}) imposes $L_1$-sparsity making use of the scaling factor from the batch normalization; later progressive pruning methods (i.e., gradually increase pruning ratio) are developed to improve the resulting models' accuracy~\cite{ye2018progressive}. Another stream of approaches involves designing compact models, such as MobileNet \cite{mobilenetv2} and ShuffleNet~\cite{zhang2018shufflenet}. Energy cost was leveraged in~\cite{yang2017designing} to guide the pruning towards the goal of energy-efficient inference.

Resource-efficient training is different from and more complicated than its inference counterpart. However, many insights gained from the latter can be lent to the former. For example, the recent work~\cite{prunetrain} showed that performing active channel pruning during training can accelerate the empirical convergence. Lately, Wang et al.~\cite{e^2_train} proposed one of the first comprehensive energy-efficient training frameworks, consisting of stochastic data dropping, selective layer updating, and low-precision back-propagation. They demonstrated its success in training several classification models with over 80\% energy savings. \cite{biggest_loser} accelerated training by skipping samples that may lead to low loss values (considered as less informative) at each iteration.

\subsection{Semantic segmentation}

\textbf{Multi-scale aggregation in segmentation.}
Multi-scale aggregation has been proven to be powerful for semantic segmentation~\cite{deeplabv3+,yu2018bisenet,zhao2018icnet,zhao2017pyramid}, via integrating multi-scale modules and high-/low-level features to capture patterns of different granularities. Pyramid Pooling and Atrous Spatial Pyramid Pooling (ASPP) modules were introduced in~\cite{zhao2017pyramid} and \cite{deeplabv3+} to aggregate features learned in different sizes of receptive fields, adapting the models to objects with different semantic sizes. Parallel branches of different downsampling rates were proposed by~\cite{yu2018bisenet,zhao2018icnet} to cover different resolutions. Although multi-scale aggregation contributes to segmentation accuracy improvement, it and its associated header introduce extra overhead during both training and inference (e.g., 52.98\% inference FLOPs of Deeplabv3+ with a ResNet50 backbone and output stride of 16). That motivates us to slim such modules in \FrameworkName.

\textbf{Efficient segmentation models.}
A handful of efficient semantic segmentation models have been developed:
ENet~\cite{paszke2016enet} used an asymmetric encoder-decoder structure together with early downsampling; ICNet~\cite{zhao2018icnet} cascaded feature maps from multi-resolution branches under proper label guidance, together with \NOTE{network} compression; and BiSeNet~\cite{yu2018bisenet} fused a context path with a fast downsampling scheme and a spatial path with smaller filter strides. 

\textbf{Remaining challenges.}
However, the models above were neither \textit{customized for} nor \textit{evaluated on} ultra-high resolution images, and our experiments show that they did not achieve sufficiently satisfactory trade-off in such cases. A knowledge distillation method was also leveraged to boost the performance of a computationally light-weight segmentation model from a teacher network~\cite{he2019knowledge}. Despite their progress, none of them touches the training efficiency, nor any discussion related to \textit{co-optimization} with the input data. Besides, the FLOPs number has a correlation to, but is not a faithful indicator of the actual energy cost, as pointed out by many prior works~\cite{yang2017designing}.
\begin{figure*}[!t]
\begin{center}
   \includegraphics[width=1.0\linewidth]{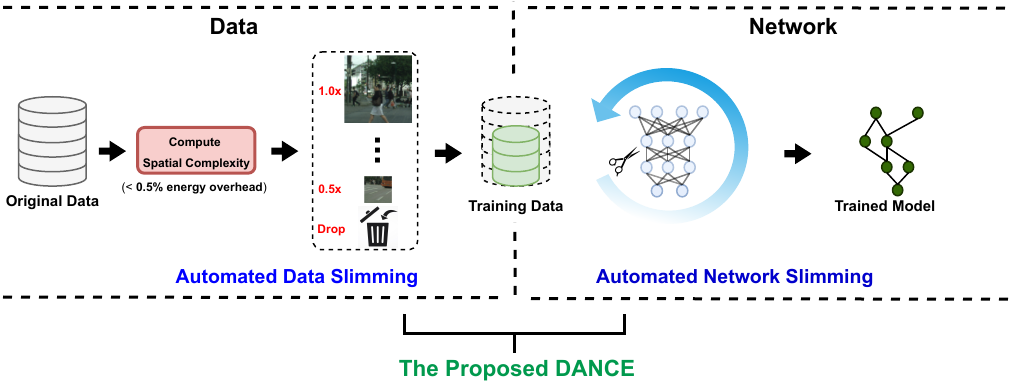}
\end{center}
\vspace{-1em}
   \caption{An overview of the data-\NOTE{network} co-optimization pipeline in the proposed \FrameworkName~framework.}
\label{fig:cls_pipeline}
\end{figure*}

\section{The proposed \FrameworkName~framework}
\label{sec:proposed}
This section presents our proposed \FrameworkName~framework. We will first provide an overview of \FrameworkName~in Section \ref{sec:overview}, and then introduce \FrameworkName's automated \textit{data} slimming and automated \textit{\NOTE{network}} slimming design in Section \ref{sec:auto-data} and Section \ref{sec:auto-model}, respectively. 

\subsection{\FrameworkName~overview}
\label{sec:overview}
The driving hypothesis of \FrameworkName~is that matching the data and \NOTE{network} can potentially boost both the model performance and hardware efficiency by removing redundancy associated with both the data and \NOTE{network}. As such,  \FrameworkName~aims to reduce the computational and energy costs of segmentation tasks during both training and inference, via \textbf{a joint effort} from \textbf{data-level} and \textbf{\NOTE{network}-level}. Specifically, as shown in Fig.~\ref{fig:cls_pipeline}, \FrameworkName~integrates both automated data and \NOTE{network} slimming, where the former automatically performs \textit{complexity-driven} data downsampling/dropping before applying the data to a \NOTE{network} while the latter automatically and progressively prunes the \NOTE{network} to \textbf{match the slimmed data}. A bonus benefit of \FrameworkName~is that the resulting data-\NOTE{network} pipeline after training (i.e., inference) is also naturally cost-efficient. 

\subsection{\FrameworkName: automated data slimming}
\label{sec:auto-data}
\FrameworkName's automated \textit{data} slimming strives to automatically downsample or drop input images and controls their corresponding contribution to the training loss, \textbf{adapting to} the images’ spatial complexity which is estimated using a spatial complexity indicator. 

\textbf{Spatial complexity indicator.}
Spatial complexity has been commonly used as the basis for estimating image complexity~\cite{gain2019relating,mishra2019cc,xu2019approxnet}, such as the one proposed in~\cite{image_complexity}:
\begin{equation}\label{SI}
    {SC_{mean} = \frac{1}{M}\sum \sqrt{s_h^2+s_v^2}}
\end{equation}

where ${s_h}$ and ${s_v}$ denote gray-scale images filtered with horizontal and vertical Sobel kernels, respectively, and ${M}$ denotes the number of pixels.
Developed by~\cite{image_complexity} to predict the image complexity for imaging compression/coding purpose, $SC_{mean}$ reflects the pixel-level variances and is extremely efficient to calculate, e.g., account for only $0.15\%$ FLOPs and $<0.5\%$ energy (on-device measurement when including both computations and data movements) of the DeepLabv3+ model (ResNet50 as the backbone with an output stride of 16) on one  RGB image patch of size 224$\times$224. 

\begin{wrapfigure}{r}{0.5\linewidth}
    \centering 
    \vspace{-2em}
  \includegraphics[width=1.0\linewidth]{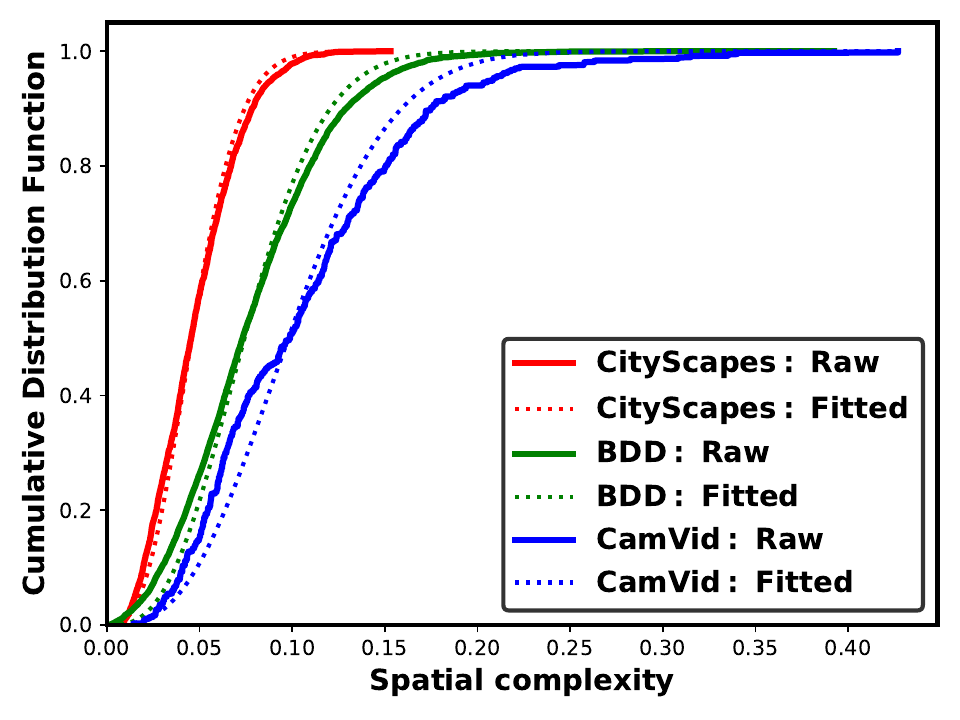}
  \vspace{-2em}
  \caption{The image complexity distribution of the training sets in the three considered urban scene understanding datasets.}
\label{fig:ic_threshold}
\end{wrapfigure}

In \FrameworkName, we first compute  all training samples' $SC_{mean}$ and fit the corresponding cumulative distribution function (CDF) using a Maxwell-Boltzmann distribution~\cite{maxwell1860v}, which turns out to be well-matched in all considered datasets as shown in Fig.~\ref{fig:ic_threshold}. Statistical analysis of $SC_{mean}$ for a specific dataset is an interesting question, which we leave for future works.

Thanks to the fitted CDF, given an input image, we can project its $SC_{mean}\in [0, inf]$ to a variable $p\in[0, 1]$ via probability integral transform~\cite{dodge2006oxford}. The resulting $p$ is then directly used as the corresponding input image's downsampling ratio, stochastic dropping probability, and weighted coefficient in the training loss.

\label{sec:downsampling}
\textbf{Complexity-adaptive downsampling.}
The proposed complexity-adaptive downsampling in \FrameworkName~draws inspiration from recent findings which show that \textit{not all input samples are born equal}~\cite{biggest_loser,Li_2017}, and is motivated by the fact that downsampling input image sizes can most straightforwardly reduce the training/inference energy costs, as well as directly benefits the memory throughput. Meanwhile, a few recent works learn to adjust resolution or respective fields~\cite{dai2017deformable,marin2019efficient}, whose promising results further motivate our complexity-adaptive downsampling. 

As prior works show that the minimal acceptable downsampling ratio is 0.5 for most segmentation models
\cite{yu2018bisenet,deeplabv3+}, we make use of the spatial complexity indicator $SC_{mean}$ to downsample the input images with a  ratio of $(0.5p + 0.5)\in[0, 0.5]$, where $p$ is the aforementioned projected value  corresponding to the images' $SC_{mean}$. In contrast to the learning-based approaches in prior works~\cite{dai2017deformable,marin2019efficient} that incur extra training workloads, we seek a reliable indicator that is mostly ``training free'' and inexpensive to compute, based on which we can estimate a proper downsampling rate per image adaptively. In particular, the energy overhead of our complexity-adaptive downsampling is <$0.02\%$ when estimated using real-device measurements in all our considered datasets.

\textbf{Complexity-adaptive stochastic dropping.}
\label{sec:casd}
Recent pioneering CNN efficient works~\cite{biggest_loser,e^2_train} proposed that dropping a portion of training samples/mini-batches, either randomly or using some loss-based deterministic rules, can reduce the total training costs without notably sacrificing or even improving the algorithmic accuracy. Inspired by the stochastic dropping idea of~\cite{e^2_train}, we incorporate the readily available spatial complexity indicator in Eq. (\ref{SI}) to calibrate the dropping probability. Specifically,~\cite{e^2_train} proposes to randomly skip incoming data (in mini-batch) with a default probability of 50\% (i.e., 50\% of the data is discarded without being fed into the models). The authors demonstrated this naively simple idea (with zero overhead) to be highly effective for efficient training without hurting or even improving the achieved accuracy. We further hypothesize that the images with larger spatial complexity are more informative and likely to favor the achieved accuracy if being more frequently trained than the ones with smaller spatial complexity. 

Therefore, instead of adopting a uniformly dropping probability for all images, we propose a simple yet effective heuristic to enable complexity-adaptive stochastic dropping by assigning a smaller dropping probability to input images with larger spatial complexity. In particular, we assign $(1-p)$ as the dropping probability, where $p$ is the aforementioned projected value of the images' spatial complexity indicator ($SC_{mean}$). 

\textbf{Complexity-adaptive loss. } 
Similarly, the losses produced by images with different complexities have been observed to contribute differently to the training loss~\cite{gain2019relating} or convergence in training~\cite{biggest_loser}. We thus prioritize the updates generated by samples with larger spatial complexity, and adopt an adaptive weighted loss as below:

\begin{equation}\label{Weighted_L}
    \mathcal{L} = \frac{\sum w_i \cdot l_i}{\sum w_i}= \frac{\sum p_i \cdot l_i}{\sum p_i}, i = 1,2,... N
\end{equation}
where $w_i$ is a scalar weighted coefficient, and $l_i$ is the cross-entropy loss of samples, corresponding to the $i$-th image of the current mini-batch with $N$ images. 
Similar to the dropping probability assignment in complexity-adaptive stochastic dropping, input image with larger spatial complexity will be assigned a larger weighted coefficient than the one with smaller spatial complexity. As such, we adopt weighted coefficients equal to the aforementioned projected value $p$ of the images' spatial complexity indicator ($SC_{mean}$), i.e., $w_i=p_i,i = 1,2,... N$.

\subsection{\FrameworkName: automated \NOTE{network} slimming}
\label{sec:auto-model} 
Various ways to aggregate multi-scale features~\cite{deeplabv3+,yu2018bisenet,zhao2018icnet,zhao2017pyramid} have been proved to improve segmentation accuracy at a cost of extra parameters and computations, leading to a higher training/inference energy burden. Thanks to the developed complexity-adaptive downsampling in \FrameworkName's automated \textit{data} slimming (see Section \ref{sec:downsampling}), the resulting inputs have been re-scaled according to their spatial complexity. We conjecture that such downsampled inputs naturally have more ``normalized'' object feature scales, i.e., complexity-adaptive downsampling can emulate the effect of multi-scale aggregation, and thus can potentially rely less on multi-scale aggregation modules for improving the segmentation accuracy. We thus expect that the \NOTE{network} appears to be more redundant when handling our automated \textit{data} slimming's resulting downsampled inputs as the cost dominant building blocks of multi-scale aggregation now becomes less important. 

\textbf{Progressive pruning during training.}
Motivated by the above conjecture and targeting reduced costs for both the training and inference (e.g., post-training pruning merely reduces inference costs), we propose an automated \NOTE{network} slimming with a progressive pruning schedule during the training trajectory  to prune the header of \NOTE{the networks for segmentation}, which includes the aforementioned multi-scale feature modules and also often dominates both the training and inference costs, e.g., accounts for 52.98\% FLOPs in DeepLabv3+ (with a ResNet50 backbone and an output stride of 16). Note that \FrameworkName's effectiveness and insights extend when other \NOTE{network} pruning methods are considered, here we consider progressive pruning without loss of generalization.

To design the progressive pruning schedule, we develop a straightforward heuristic design, following the commonly used schedule in most pruning works \cite{Renda2020Comparing,luo2017thinet,han2015learning}. Specially, we first divide the whole training/adaptation process into several stages w.r.t the total number of iterations, and then perform channel-wise pruning (based on~\cite{liu2017learning}) at the end of each stage.

\begin{figure}[t]
\begin{center}
   \includegraphics[width=0.9\linewidth]{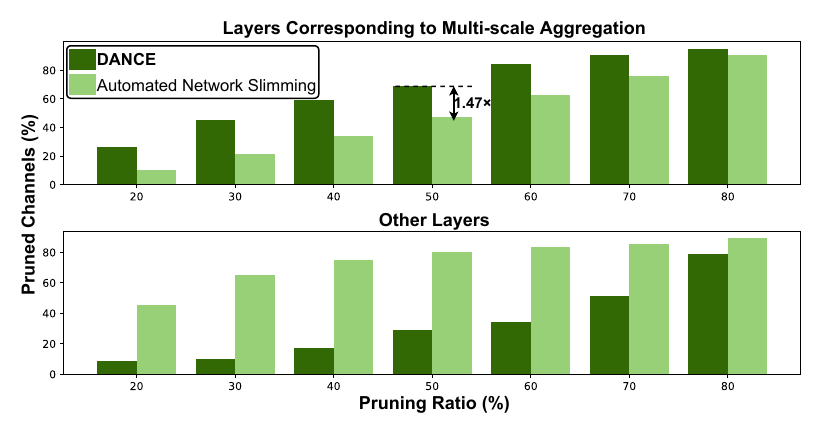}
\end{center}
\vspace{-1em}
   \caption{The percentage of pruned channels for different kinds of layers under various pruning ratios on a DeepLabv3+ model (with a ResNet50 backbone and an output stride of 16) with the Cityscapes dataset.}
\label{fig:prune}
\end{figure}

\textbf{Co-optimization affects pruning patterns.}

To validate the aforementioned conjecture, we visualize the percentage of pruned channels in layers corresponding to multi-scale aggregation and other layers under different pruning ratios in Fig.~\ref{fig:prune}, when the models are trained with \FrameworkName~or merely \FrameworkName's automated \NOTE{network} slimming.

We can see that training with both automated data and \NOTE{network} slimming, i.e., \FrameworkName, always prunes more channels in layers corresponding to multi-scale aggregation (e.g., the ASPP module in DeepLabv3+) and fewer channels on other layers, under all the considered seven pruning ratios between 20\% and 80\%, while merely automated \NOTE{network} slimming does opposite. Specifically, as compared to training using merely automated data slimming under the same pruning ratio of 50\%, the model trained with both automated data and \NOTE{network} slimming, i.e., models trained using \FrameworkName, prunes 1.47$\times$ more channels in layers associated with multi-scale aggregation, where the corresponding accuracy is also higher (e.g., a 5.33\% higher mIoU on the Cityscapes validation dataset together with a 54.8\% lower inference energy with images of 592 $\times$ 592).

The experiment in Fig.~\ref{fig:prune} together with those in the experiment section verify our conjecture that (1) matching the data with the \NOTE{network} can potentially improve the accuracy (thanks to the match between slimmed data and unpruned channels' distribution) and remove redundant costs associated with both the data and \NOTE{network}, thus achieving ``all-win'': reducing both the training and inference costs while improving the achieved model accuracy (mIOU); and (2) \FrameworkName's automated data slimming can (partially) emulate the effect of multi-scale aggregation in segmentation models, enabling a higher pruning ratio on the corresponding multi-scale aggregation modules. The observations are consistent when other pruning methods \NOTE{and different pruning hyperparameters} are used in~\FrameworkName's automated \NOTE{network} slimming (more details in Section~\ref{sec:unstructure_pruning}), again verifying that the above conclusion (i.e., ``co-optimization affects the optimal pruning patterns") holds for~\FrameworkName~regardless of the adopted pruning designs.

\section{Experiments}
\label{sec:exp}
In this section, we evaluate \FrameworkName~on four segmentation models and three popular urban scene understanding datasets in terms of mIoU and the total training/inference FLOPs and energy cost, where the energy cost is measured when training/inference the corresponding models in a SOTA edge device (JETSON TX2~\cite{edgegpu}).
We consider both the computational and energy costs because the former is commonly adopted and thus helps to benchmark with prior works while the latter better capture the real hardware cost. 

\subsection{Experiment setting}

\textbf{Considered models and datasets.} Our evaluation of \FrameworkName~considers four SOTA segmentation models (two complicated models: DeepLabv3+~\cite{deeplabv3+}, PSPNet~\cite{zhao2017pyramid}, and two compact models: ICNet \cite{zhao2018icnet}, and BiSeNet~\cite{yu2018bisenet}) and three commonly used urban scene understanding datasets (Cityscapes~\cite{cityscapes}, CamVid~\cite{brostow2008segmentation}, and BDD~\cite{yu2018bdd100k}) in many efficient segmentation models~\cite{chen2019fasterseg,zhao2018icnet,yu2018bisenet}.

\textbf{Experimental platforms and training details.} All experiments (except the energy measurements) are performed on a workstation with NVIDIA 2080Ti GPU cards using the PyTorch framework \cite{paszke2017automatic} for a fair comparison. We use an SGD optimizer with a learning rate of $1\times10^{-3}$ for training all models except ICNet, which adopts a learning rate of $1\times10^{-2}$ due to the unavailability of the corresponding ImageNet pre-trained model; and a minibatch size of (1) 8 for the DeepLabv3+ and PSPNet models and (2) 16 for the BiSeNet and ICNet models.

\begin{wraptable}{r}{0.52\linewidth}
\vspace{-1.5em}
\caption{The FLOPs, energy cost, and mIoU of \FrameworkName~on top of the four models on the \textbf{Ciytscapes} test dataset.}
\label{fig:ic_threshold}
  \resizebox{1\linewidth}{!}
  {    
    \begin{tabular}{c||ccccc}
    \toprule
    \multirow{2}{*}{\textbf{Model}} & \multicolumn{2}{c}{\textbf{FLOPs}} & \multicolumn{2}{c}{\textbf{Energy}} & \textbf{mIoU} \\
     & Train. (P) & Infer. (G) & Train. (MJ)& Infer. (J) & (\%) \\ \midrule
    DeepLabv3+ & 198.31 & 743.64 & 45.21  & {\color{black}41.32}& 68.05 \\
    \textbf{\FrameworkName~Improv.} & \textbf{-35.75\%}  & \textbf{-53.67\%} & \textbf{-39.69\%} & \textbf{-45.02\%} &  \textbf{+3.44} \\ \midrule
    PSPNet &  153.61 &  582.54 &  35.16 & 30.99 &  65.59 \\
    \textbf{\FrameworkName~Improv.}  &  \textbf{-39.28\%} & \textbf{-50.23\%}  & \textbf{-34.92\%} & \textbf{-42.81\%}  & \textbf{+1.93} \\ \midrule
    ICNet & 39.33 & 45.20 & 10.82  & 6.51 & 47.74 \\
    \textbf{\FrameworkName~Improv.} & \textbf{-49.77\%}  & \textbf{-55.67\%}  &  \textbf{-45.27\%} & \textbf{-47.69\%} & \textbf{+13.34} \\ \midrule
    BiSeNet & 73.64 & 157.41  & 18.40 & 9.93 &  71.69 \\
    \textbf{\FrameworkName~Improv.}  & \textbf{-32.77\%}  &  \textbf{-39.29\%} & \textbf{-25.66\%} &  \textbf{-31.27\%} &  \textbf{-0.71}\\ \bottomrule
    \end{tabular}
    }
  \label{table:res_cityscapes}
  \vspace{-1.5em}
\end{wraptable}

\subsection{Performance on various datasets/models}
In this subsection, we apply \FrameworkName~to the four segmentation models and three datasets and compare the resulting segmentation accuracies and inference/training costs with those of the base models. 

\subsubsection{\FrameworkName~on the Cityscapes dataset}
Table \ref{table:res_cityscapes} compares the segmentation accuracy, and computational and energy costs of \FrameworkName~on the four models, i.e., DeepLabv3+~\cite{deeplabv3+}, PSPNet~\cite{zhao2017pyramid}, ICNet~\cite{zhao2018icnet}, and BiSeNet~\cite{yu2018bisenet}, when evaluated on the Cityscapes dataset. We can see that (1) \FrameworkName~saves about 36\% - 39\% and 35\% - 40\% computational and energy costs in training (a similar trend in inference), while boosting the mIoU in the cases of DeepLabv3+~\cite{deeplabv3+} and PSPNet~\cite{zhao2017pyramid} by 3.44\% and 1.93\%, respectively; (2) In the case of ICNet, \FrameworkName~achieves a 13.34\% higher mIoU with up to 45\% energy savings than those of the base model, where the lower mIoU of the base model might be due to the lack of a corresponding ImageNet pre-trained model; and (3) Though \FrameworkName~doesn't boost the mIoU on the compact model of BiSeNet, it does save in training energy cost and win bigger (saving up to 31\% energy) in inference.

\begin{wraptable}{r}{0.52\linewidth}
\vspace{-1.5em}
\caption{The FLOPs, energy cost, and mIoU of \FrameworkName~on top of the four models on the \textbf{CamVid} test set.}
\label{fig:ic_threshold}
  \resizebox{1.0\linewidth}{!}
  {    
     \begin{tabular}{c||ccccc}
    \toprule
    \multirow{2}{*}{\textbf{Model}} & \multicolumn{2}{c}{\textbf{FLOPs}} & \multicolumn{2}{c}{\textbf{Energy}} & \textbf{mIoU} \\
     & Train. (P) & Infer. (G) & Train. (MJ) & Infer. (J) & (\%) \\ \midrule
    DeepLabv3+& 37.19 & 254.62 & 7.30 & 20.19  &69.15 \\
    \textbf{\FrameworkName~Improv.} & \textbf{-32.76\%} & \textbf{-47.6\%} & \textbf{-31.76\%} & \textbf{-43.65\%} &  \textbf{+1.51} \\ \midrule
    PSPNet &  27.27 & 208.77  &  4.71 & 14.64 & 65.28 \\
    \textbf{\FrameworkName~Improv.} & \textbf{-39.69\%} &  \textbf{-46.47\%} & \textbf{-32.22\%} & \textbf{-41.22\%}  & \textbf{+2.82} \\ \midrule
    ICNet & 6.01 & 16.33 & 1.78 & 4.21 & 53.29 \\
    \textbf{\FrameworkName~Improv.} &  \textbf{-45.32\%} & \textbf{-52.64\%} & \textbf{-49.21\%} & \textbf{-56.21\%}  & \textbf{+1.40} \\ \midrule
    BiSeNet & 6.46 & 54.17 & 2.72  &  6.77  & 68.6 \\
    \textbf{\FrameworkName~Improv.} & \textbf{-38.09\%} & \textbf{-41.10\%}  & \textbf{-32.45\%} & \textbf{-33.76\%} & \textbf{-0.27} \\ \bottomrule
    \end{tabular}
    }
\vspace{-1.5em}
  \label{table:res_camvid}
\end{wraptable}

\subsubsection{\FrameworkName~on the CamVid dataset}
Under smaller images ($720\times960$) in CamVid (vs. $1048\times2048$ in Cityscapes), we can still observe similar trends as those in Cityscapes (see Table \ref{table:res_cityscapes}). Specifically, 
our \FrameworkName~can still save 32\% - 49\% energy cost, as shown in Table \ref{table:res_camvid}, while achieving improved mIoU (over 1.4\%). For the compact model BiSeNet, with a comparable mIoU, our \FrameworkName~still stably brings 32\% and 33\% energy savings in training and inference, respectively.

\subsubsection{\FrameworkName~on the BDD dataset for adaptation}
As Section \ref{sec:intro} stated, for most on-device learning applications, training from scratch is not necessary and the ability to adapt to new data can be more interesting for some applications, especially for autonomous vehicles and robots.

\begin{wraptable}{r}{0.52\linewidth}
\caption{The FLOPs, energy cost, and mIoU of \FrameworkName~on top of the four models on the \textbf{BDD} test set on adaptation.}
\label{fig:ic_threshold}
  \resizebox{1.0\linewidth}{!}
  {    
\begin{tabular}{c||ccccc}
    \toprule
    \multirow{2}{*}{\textbf{Model}} & \multicolumn{2}{c}{\textbf{FLOPs}} & 
    \multicolumn{2}{c}{\textbf{Energy}} &
    \textbf{mIoU} \\
     & Train. (P) & Infer. (G) & Train. (MJ) & Infer. (J) & (\%) \\ \midrule
    DeepLabv3+& 97.01 & 339.46 & 17.74 & {\color{black}27.67} & 52.66 \\
    \textbf{\FrameworkName~Improv.} & \textbf{-79.31\%} & \textbf{-51.86\%} & \textbf{-77.15\%} & \textbf{-43.5\%} & \textbf{+0.12} \\ \midrule
    PSPNet & 72.56 & 290.57  &  11.82  &  19.37 & 39.54 \\
    \textbf{\FrameworkName~Improv.} & \textbf{-37.47\%} & \textbf{-49.65\%} & \textbf{-25.18\%} & \textbf{-40.29\%}& \textbf{+5.51} \\ \midrule
    ICNet & 58.27 & 21.68 & 15.73 & {\color{black}5.17 }& 39.53 \\
    \textbf{\FrameworkName~Improv.} & \textbf{-58.65\%} & \textbf{-51.51\%} & \textbf{-61.47\%} & \textbf{-37.47\%} & \textbf{+0.47} \\ \midrule
    BiSeNet & 45.32 & 72.27 & 7.48 & {\color{black}7.94} & 56.20 \\
    \textbf{\FrameworkName~Improv.} &  \textbf{-34.08\%} & \textbf{-44.47\%} & \textbf{-27.92\%} & \textbf{-38.46\% }& \textbf{ +0.27} \\ \bottomrule
    \end{tabular}
    }
  \label{table:res_bdd}
\end{wraptable}

Here, we choose the BDD~\cite{yu2018bdd100k} for the adaptation experiments. We use pre-trained models on Cityscapes to adapt to unseen images in BBD. For a fair comparison, we choose the same checkpoints as the pre-trained model for each model in experiments. The adaptation performance
is summarized in Table \ref{table:res_bdd}, which shows that while being similar to the  performance on Cityscapes, \FrameworkName~saves up to 77\% energy cost while achieving a slightly better (+0.12\%) mIoU over the baseline, or boosts the mIoU by 5.51\% when requiring even a 25\% lower energy cost than the baseline. 

The extensive results in Tables \ref{table:res_cityscapes} - \ref{table:res_bdd} show that \textbf{\FrameworkName~can achieve ``all-win'' on all the three datasets when applying to both DeepLabv3+ and PSPNet}: lower training cost (energy savings: 77\% - 25\%), more efficient inference (energy savings: 40\% - 45\%), and improved mIOU (0.12\% - 5.51\%), demonstrating \textbf{the consistent superiority of \FrameworkName~on complicated models}. As for the performance on compact models, \FrameworkName~can improve efficiency of both training (energy savings: 25\% - 61\%) and inference (energy savings: 31\% - 56\%) with a slightly dropped or even better mIoU (-0.71\% - 13.34\%) on all the three datasets, indicating that \textbf{\FrameworkName~can benefit energy efficiency of even compact models}.

\subsection{Ablation studies of \FrameworkName}
\label{sec:abalation}

\begin{figure}[b]
\vspace{-1em}
\begin{center}
   \includegraphics[width=0.7\linewidth]{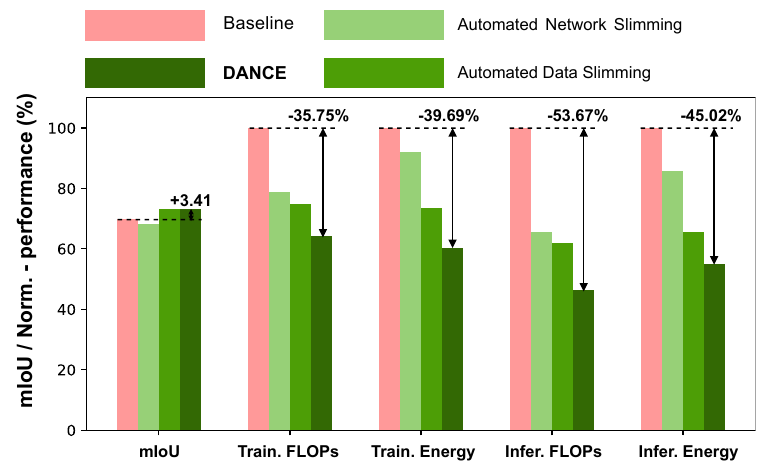}
\end{center}
\vspace{-1em}
   \caption{Ablation studies of data-\NOTE{network} co-optimization in \FrameworkName~on the DeepLabv3+ model (with a ResNet50 backbone and an output stride  of 16) and Cityscapes validation dataset, where FLOPs and energy are normalized to those of the baseline.}
\label{fig:abl}
\end{figure}
In this subsection, we perform ablation studies of \FrameworkName~for evaluating the effectiveness of its data-\NOTE{network} co-optimization, $p$ indicator, and automated data slimming.

\subsubsection{Ablation study on the effectiveness of~\FrameworkName's data-\NOTE{network} co-optimization}

\begin{wraptable}{r}{0.5\linewidth}
\caption{The FLOPs and mIoU of \textbf{co-optimize} and \textbf{separately optimize} on top of DeepLabv3+@CityScapes.}
\label{table:seqVSco}
\centering
\resizebox{1.0\linewidth}{!}
{
\begin{tabular}{c||ccc}
\toprule
\multirow{2}{*}{\textbf{Method}} & \multicolumn{2}{c}{\textbf{FLOPs}} & 
\textbf{mIoU}  \\
 & Train. (P) & Infer. (G) & (\%) \\
\midrule
Baseline  & 198.31 & 743.64 & 69.71 \\
\midrule
Optimize \NOTE{Network} After Data  & +0.85\% & -61.10\% & +0.66 \\
\midrule
Optimize Data After \NOTE{Network}  & +12.99\% & -55.88\% & +3.08 \\
\midrule
\textbf{Co-Optimize (\FrameworkName)} & \textbf{-35.75\%} & \textbf{-56.57\%} & \textbf{+3.41} \\
\bottomrule
\end{tabular}
}
\vspace{-1em}
\end{wraptable}

\textbf{\FrameworkName~ vs. only automated \NOTE{network}/data slimming.} As shown in Fig.~\ref{fig:abl}, combining both automated data and \NOTE{network} slimming (i.e., \FrameworkName) achieves (1) better performance (in terms of training cost, inference cost, and mIoU) than the standalone implementation of either of these two techniques integrated into \FrameworkName~(i.e., automated data and \NOTE{network} slimming); and (2) a much higher mIoU than the baseline (+3.41\%) while requiring 39\% and 45\% less energy in training and inference, respectively.
This set of experiments indicates the advantage of jointly matching the data and \NOTE{network} for co-optimization.

\noindent\textbf{\FrameworkName~ vs. optimizing \NOTE{network} and data separately.} 
Table~\ref{table:seqVSco} compares \textbf{co-optimization} (\FrameworkName) with \textbf{separate optimization} (optimizing \NOTE{network}/data and then data/\NOTE{network} sequentially), showing that \NOTE{network} and data need to be jointly co-optimized to achieve the best mIoU-cost trade-off, while optimizing (i.e., slimming) the \NOTE{network} and data sequentially will cause a 0.33\% - 2.75\% mIoU drop on DeepLabv3+@Cityscapes at an even higher computational cost (e.g., +48.74\%) than ~\FrameworkName.

\subsubsection{Ablation study of~\FrameworkName on objects with different scales}
\begin{wraptable}{r}{0.52\linewidth}
\vspace{-1.5em}
\caption{The inference mIoU of w/o \FrameworkName~and w/ \FrameworkName~on top of DeepLabv3+ for CityScapes's large, medium, and small scale of objects (manually picked \textbf{static} scales), where \FrameworkName~can further provide \textbf{dynamic} scales.}
\resizebox{1\linewidth}{!}
{
\begin{tabular}{c||c|c|c|c}
\hline
Method & \multicolumn{3}{c|}{w/o \FrameworkName}  & \textbf{w/ \FrameworkName}  \\
\hline
Image Scales  & 368$\times$368 & 496$\times$496 & 592$\times$592 & \textbf{Dynamic}  \\
\hline
IoU of Wall (\%)  & \textbf{50.56} & 48.68 & 46.40 & \textbf{52.69} \\
\hline
IoU of Motorcycle (\%)  & 53.96 & \textbf{57.51} & 55.09 & \textbf{58.23} \\
\hline
IoU of Traffic Sign (\%)  & 70.17 & 72.41 & \textbf{72.94} & \textbf{73.58} \\ \hline
\end{tabular}
}
\vspace{-1.em}
\label{table:different_scales}
\end{wraptable}

Here we compare the inference mIoU when turning off and on our \FrameworkName~applied on top of DeepLabv3+, when testing
representative large, medium, and small scales (i.e., wall, motorcycle, and traffic sign) of objects in Cityscapes. As shown in Table~\ref{table:different_scales}, we can see that (1) small/large scales of objectives favor/degrade the achieved inference mIoU of applying DeepLabv3+ to the selected objects of different scales; and (2) \FrameworkName, which inherently incorporates \textbf{dynamic} scales to its applied data, consistently outperforms its baselines even for the manually selected objectives which have \textbf{static} scales by design, indicating the advantage of \FrameworkName's automated choices of adaptive scales of data, validating \FrameworkName's inherent advantages in handling datasets/tasks of which the objects have different scales
, which is common for semantic segmentation datasets (e.g., Cityscapes~\cite{cityscapes}, CamVid~\cite{brostow2008segmentation}, and BBD~\cite{yu2018bdd100k}).

\begin{wraptable}{r}{0.52\linewidth}
\vspace{-1.5em}
\caption{The mIoU of using \textbf{proposed $p$}  (in Section~\ref{sec:auto-data}), \textbf{random $p$}, or \textbf{inverse $p$} indicator in \FrameworkName~on top of DeepLabv3+@CityScapes under same training cost budget.}
\label{table:p_indicator}
\centering
\resizebox{1.0\linewidth}{!}
{
\begin{tabular}{c||ccc}
\toprule
\textbf{Method} & \textbf{Train. FLOPs (P)} & \textbf{mIoU(\%)} \\
\midrule
\textbf{Proposed $p$} indicator  & \textbf{127.41} & \textbf{73.12} \\
\midrule
\textbf{Random $p$} indicator & +0.06\% & -4.08  \\
\midrule
\textbf{Inverse $p$} indicator, i.e., 1-$p$  & +0.00\% & -11.03  \\
\bottomrule
\end{tabular}
\vspace{-1.5em}
}
\end{wraptable}

\subsubsection{Ablation study of the $p$ indicator' effectiveness}

\begin{figure*}[t]
\centering
\vspace{-1em}
  \includegraphics[width=1.0\linewidth]{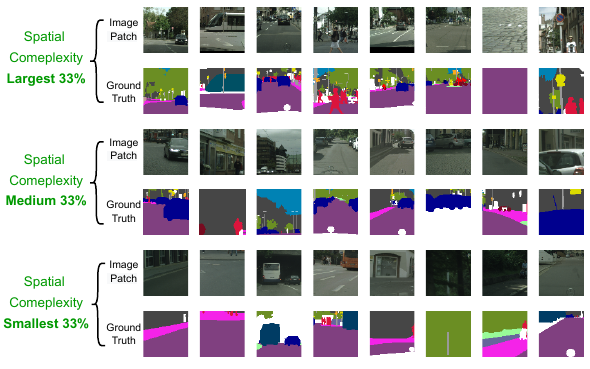}
\vspace{-2em}
  \caption{Visualizing the images randomly selected from image groups with the \textbf{largest} 33\%, \textbf{medium} 33\%, and \textbf{smallest} 33\% spatial complexity in the Cityscapes \cite{cityscapes} training dataset.}
\label{fig:picked_samples}
\vspace{-1.5em}
\end{figure*}

The spatial complexity indicator presented in Section~\ref{sec:auto-data} is to provide a variable $p\in[0, 1]$ for estimating a given image's complexity, which will be directly used to guide the slimming direction (e.g., image's downsampling ratio). As shown in Table~\ref{table:p_indicator}, we apply \textbf{inverse $p$} or \textbf{random $p$} to replace the \textbf{proposed $p$} indicator in ~\FrameworkName, and find that their resulting mIoU drops 11.03\% or 4.08\% under the same training cost budget, respectively, validating the advantageous effectiveness of our proposed $p$ indicator. Additionally, Fig.~\ref{fig:picked_samples} visualizes 24 image samples randomly selected from the image groups with the \textbf{largest} 33\%, \textbf{medium} 3\%, and \textbf{smallest} 33\% spatial complexity in the Cityscapes \cite{cityscapes} training dataset. Interestingly, we can see as expected that the image complexity identified by the adopted indicator is consistent with that by human eyes, e.g., images with spatial complexity falling within the smallest 33\% of the dataset have a simpler background and include fewer objectives. 

\subsubsection{Ablation study of \FrameworkName's effectiveness regardless of the adopted pruning methods}
\label{sec:unstructure_pruning}
\begin{wraptable}{r}{0.42\linewidth}
\vspace{-1.5em}

\caption{The number of pruned weights for different kinds of layers under various pruning ratio on a DeepLabv3+ model (with a ResNet50 backbone and an output stride of 16) with the Cityscapes dataset)}
\label{table:unstructure_pruning}
\centering
\resizebox{0.9\linewidth}{!}
{
\begin{tabular}{c||cc}
\toprule
\textbf{Pruning}& \multicolumn{2}{c}{\textbf{\#pruned weights (\FrameworkName's - AMS's)}} \\
\cmidrule{2-3}
\textbf{Ratio} & ASPP Module Layers & Other Layers \\
\midrule
20\% & 61759 & -61759 \\
\midrule
30\% & 96455 & -96455 \\
\midrule
40\% & 102535 & -102535 \\
\midrule
50\% & 96674 & -96674 \\
\midrule
60\% & 76767 & -76767 \\
\midrule
70\% & 51177 & -51177 \\
\midrule
80\% & 24986 & -24987 \\
\bottomrule
\end{tabular}
}
\vspace{-1.5em}
\end{wraptable}

We consistently find that \FrameworkName's advantages in enabling data model co-optimization is effectiveness regardless of the adopted pruning methods. For example, Table~\ref{table:unstructure_pruning} summarizes the pruning results when turning on and off \FrameworkName's automated data slimming during pruning, where we adopt the unstructured pruning in ~\cite{han2015deep}. Again, similar observations can be made as those in ~\cite{liu2017learning} when using channel-wise pruning. Specifically, training with both automated data and model slimming, i.e., \FrameworkName, always prunes more weights in layers corresponding to multi-scale aggregation (e.g., the ASPP module in DeepLabv3+) and fewer weights on other layer, under all the considered seven pruning ratios between 20\% and 80\%, whereas merely using \FrameworkName's automated model slimming (AMS) does the opposite. This set of experiment results further confirm that (1) \FrameworkName's automated data slimming can (partially) emulate the effect of multi-scale aggregation in segmentation models, and thus enable a higher pruning ratio on the corresponding multi-scale aggregation modules, and (2) matching the data with model can potentially improve the model accuracy and remove redundant costs associated with both the data and model, thus achieving ``all-win'', which is consistent with the results in Fig.~\ref{fig:acc_energy}.

\subsubsection{Ablation study of~\FrameworkName's automated data slimming}

As described in Section~\ref{sec:auto-data}, \FrameworkName's \textit{automated data slimming} integrates three techniques, including complexity-adaptive downsampling (CAD), complexity-adaptive stochastic dropping (CASD), and complexity-adaptive loss (CAL), which are guided by the adopted spatial complexity indicator. In this subsection, we evaluate the efficacy of these techniques and their different combinations on top of \FrameworkName's \textit{automated \NOTE{network} slimming} (ANS) (see Section~\ref{sec:auto-model}), in terms of the resulting task accuracy (mIoU), and computational and energy savings of both inference and training, as summarized in Table \ref{table:ads_ablation}. Note that all the task accuracy and computational and energy savings are normalized to those of the standard DeepLabv3+ \cite{deeplabv3+} model and Cityscapes dataset (See row No. 1 of Table \ref{table:ads_ablation}). We next discuss the observations in terms of the \textbf{``all-win''} goal (i.e., reducing both the training and inference costs while improving the achieved model accuracy (mIoU)):

\begin{table*}[b]
\caption{Ablation studies on the component techniques of \FrameworkName's automated data slimming on the DeepLabv3+ model (with a ResNet50 backbone and an output stride  of 16) and the Cityscapes validation dataset, where  $\mathbf{\dagger}$ (i.e., No.7) is our \FrameworkName~setting.}
\begin{threeparttable}
\centering
\resizebox{1\textwidth}{!}{
{
\begin{tabular}{cccccc||ccccc} 
\toprule
\textbf{No.} & \textbf{ANS}\tnote{a}~ & \textbf{CAD}\tnote{b}~ & \textbf{CAL}\tnote{c}~ & \textbf{CASD}\tnote{d}~ & \textbf{RD}\tnote{e}~ &  \textbf{Train. FLOPs} & \textbf{Train. Energy} & \textbf{Infer. FLOPs} & \textbf{Infer. Energy} & \textbf{mIoU} \\ \midrule
1&  &  &  &  &  & 198.3 (P) & 45.21 (MJ) &743.6 (G)  & 41.32 (J) & 69.71(\%) \\
2& \checkmark &   &  &  & & -21.16\% & -8.08\% & -34.34\% & -14.36\% & -1.52 \\

3& \checkmark & \checkmark &  &   & & -57.05\% & -51.12\% & -60.96\% &-46.52\% &-2.54 \\

4& \checkmark & \checkmark & \checkmark &    & & -56.98\% & -51.45\% &  -61.15\% &-46.96\%& -0.63  \\

5& \checkmark & &  &    & \checkmark & -13.73\% & -15.90\% & -25.07\%  &-14.21\%& +1.96 \\

6& \checkmark & &  &  \checkmark  &  &  -13.92\% & -14.10\% & -25.21\% &-15.71\%& +3.33 \\

$\mathbf{7^\dagger}$ & \textbf{\checkmark} & \textbf{\checkmark} & \textbf{\checkmark} & \textbf{\checkmark}  & &  \textbf{-35.75\%} & \textbf{-39.69\%} & \textbf{-53.67\%} & \textbf{-45.02\%} & \textbf{+3.41}\\ 
 \bottomrule 
\end{tabular}
}
}
 \begin{tablenotes}[para] 
    \footnotesize
    \item[a] ANS: Automated \NOTE{Network} Slimming
    \item[b] CAD: Complexity-Adaptive Downsampling
    \item[c] CASD: Complexity-Adaptive Stochastic Dropping
    \item[d] CAL: Complexity-Adaptive Loss
    \item[e] RD: Randomly Drop 50\% \cite{e^2_train}
    \end{tablenotes}
    \end{threeparttable}
\label{table:ads_ablation}
\end{table*}

1. \ul{Complexity-Adaptive Downsampling (CAD)}: Comparing the results in Rows No. 2 and No. 3 shows that CAD+ANS (see Row No. 3, i.e., applying CAD, which has the advantage of ``training free", on top of \FrameworkName's automated \NOTE{network} slimming (ANS)), can save 42.92\% and 32.16\%  energy cost in training and inference, respectively, whereas decreasing the mIoU by 1.02\% (i.e., -1.52\% vs. -2.54\%), as compared to merely performing ANS (see Row No. 2), indicating that CAD offers a new trade-off between the achieved energy efficiency and mIoU.

2. \ul{Complexity-Adaptive Loss (CAL)}: Comparing the results in Rows No. 3, and No. 4 shows that CAL+CAD+ANS (see Row No. 4, i.e., applying CAL on top of ANS and CAD) 
can boost the mIoU by 1.91\% as compared to merely combining CAD and ANS (i.e., CAD+ANS in Row No. 3), while still reducing 51.45\% and 46.96\% energy cost in training and inference, respectively, as compared to the DeepLabv3+ baseline (Row No. 1), indicating that adding CAL on top of CAD and ANS can further boost the model accuracy while keeping the achieved energy efficiency.

3. \ul{Complexity-Adaptive Stochastic Dropping (CASD)}: First, comparing the results in Rows No. 5 and No. 6 shows that the proposed CASD (Row No. 6) can achieve a 1.37\% higher mIoU than the random dropping technique in \cite{e^2_train} (Row No. 5) under the same energy cost of both training and inference, indicating the advantage of \textit{complexity-adaptive} stochastic dropping over \textit{random} dropping in \cite{e^2_train}. Second, comparing the results in Rows No. 4, and No. 7 shows that applying CASD on top of CAL+CAD+ANS (Row No. 4) can boost the mIoU by 4.04\% as compared to merely combining ANS, CAD, and CAL (Row No. 4), and by 3.41\% as compared to the DeepLabv3+ baseline (Row No. 1), while obtaining 39.69\% and 45.02\% energy savings in training and inference, respectively, as compared to the DeepLabv3+ baseline (Row No. 1). 

This set of comparisons indicates the effectiveness of the proposed \FrameworkName's automated data slimming, i.e., integrating all three component techniques of \FrameworkName's automated data slimming can achieve the most favorable data-\NOTE{network} co-optimization benefits as it achieves the \textbf{``all-win''} goal as shown in Fig.~\ref{fig:acc_energy}.

\section{Ablation study of \FrameworkName’s hyperparameters}
\label{sec:abl_hyperparams}
In this subsection, we perform experiments for evaluating \FrameworkName~ with different hyperparameters by changing the ranges of (1) the weighted coefficients in complexity-adaptive loss (CAL) and (2) dropping probability in complexity-adaptive stochastic dropping (CAL) (as described in Section~\ref{sec:auto-data}), and summarize the results in Table \ref{table:hyperparameters}. To better study the effect of each of the aforementioned hyperparameters, we fix others with the default ones (as described in Section~\ref{sec:auto-data}) when tuning one of them. 

Note that the larger the \textit{ratio of endpoints in the dynamic range} of both the weighted coefficients and dropping probabilities are, the more (less) frequent images with a higher (lower) spatial complexity would be used. And the largest ratio is $1.0/0.0=\infty$ and $100\%/0\%=\infty$ for the weighted coefficients and dropping probabilities, respectively, which is also the default setting as mentioned in Section~\ref{sec:auto-data}.

The results in Table \ref{table:hyperparameters} show that \textbf{increasing the frequency of training images with a higher spatial complexity} (defined in Eq.~\ref{SI}), by increasing the \textit{ratio of endpoints in the dynamic range} of the weighted coefficient or dropping probability, \textbf{favors the segmentation accuracy (i.e., a higher mIoU)}. This observation is consistent with that of \cite{biggest_loser,gain2019relating}. Specifically, changing the dropping probability range from 60\% - 40\% to 100\% - 0\% boosts the achieved mIoU by 1.99\%, while changing the weighted coefficient range from 2.0 - 1.0 to 1.0 - 0.0 leads to an improved mIoU of 0.82\%, while the training and inference costs of both cases mostly stay the same.

\begin{table*}[t]
\caption{Ablation study of \FrameworkName's hyperparameters: \FrameworkName~with different ranges of (1) weighted coefficients in complexity-adaptive loss (CAL) and (2) dropping probability in complexity-adaptive stochastic dropping (CASD) on  DeepLabv3+ with Cityscapes.}

\centering
\resizebox{1\textwidth}{!}{
{
\begin{tabular}{c|c||ccccc}
\toprule
\multicolumn{2}{c||}{Settings} & Train. FLOPs & Train. Energy &Infer. FLOPs & Infer. Energy & mIoU \\ \midrule
\multirow{3}{*}{Range of the Weighted Coefficients in CAL} & 2.0 - 1.0 & 124.66 (P) & 26.43 (MJ) & 358.39 (G) & 23.57 (J) &  72.30\% \\
 & 4.0 - 1.0 & -1.76\% & -2.65\% & +0.78\% & +0.52\% & +0.64\% \\
 & 1.0 - 0.0 & +2.21\% & +3.18\%  & -3.87\% & -3.60\% & +0.82\% \\ 
\midrule
\multirow{3}{*}{Range of the Dropping Probability in CASD} &
 60\% - 40\% & 128.05 (P) & 27.55 (MJ) & 345.21 (G) & 22.74 (J)& 71.13\% \\
 & 75\% - 25\% & -0.88\% & -0.57\% &  -1.37\% & -1.22\% & +0.27\% \\
 & 100\% -0\% & -0.50\% & -1.02\%  & -0.20\% & -0.07\% & +1.99\% \\ 
\bottomrule
\end{tabular}
\vspace{-2em}
}
}\label{table:hyperparameters}
\end{table*}
\section{Conclusions}
We proposed \FrameworkName~for boosting segmentation efficiency during both training and inference, leveraging the hypothesis that maximum model accuracy and efficiency should be achieved when the data and model are optimally matched. 
On the ``data-level'', \FrameworkName's automated data slimming not only halve the computational and energy costs, but also boost the segmentation accuracy. Interestingly, \FrameworkName's automated data slimming can emulate the effect of multi-scale feature extraction yet at a much lower cost. This further motivates \FrameworkName's automated network slimming on the ``model-level'' that advocates automatically pruning the model adapting to the resulting data slimmed by \FrameworkName's automated data slimming and leads to more pruning in the cost-dominant building blocks for multi-scale feature extraction, validating our hypothesis and further reducing both training and inference costs. Extensive experiments and ablation studies validate \FrameworkName's effectiveness and superiority, which resides in its capability to automatically match the data and network via automated co-optimization.


\bibliographystyle{ACM-Reference-Format}
\bibliography{sample-base}


\end{document}